\documentclass{article}
\usepackage{ijcai19}

\usepackage{times}
\usepackage{soul}
\usepackage{url}
\usepackage[utf8]{inputenc}
\usepackage[small]{caption}
\usepackage{graphicx}
\usepackage{amssymb,amsmath,amsthm}
\usepackage{booktabs}
\usepackage{algorithm}
\usepackage{algorithmic}
\newtheorem{theorem}{Theorem}

\usepackage{xcolor}
\usepackage{colortbl}
\colorlet{tableheadcolor}{gray!25} 
\colorlet{tablerowcolor}{gray!20} 
\newcommand{\rowcol}{\rowcolor{tablerowcolor}} %

\newcommand{\NP}{\textsc{NP}}

\title{Learning Multi-Stage Sparsification for Maximum Clique Enumeration\footnote{Part of this work was done while the authors were at Nokia Bell Labs, Ireland}}

\author{
Marco Grassia$^1$\and
Juho Lauri$^2$\and
Sourav Dutta$^3$\And
Deepak Ajwani$^4$
\affiliations
$^1$University of Catania, Italy\\
$^2$Nokia Bell Labs, Ireland\\
$^3$Eaton Corp., Ireland\\
$^4$University College Dublin, Ireland
\emails
marco.grassia@studium.unict.it,
juho.lauri@gmail.com,
souravdutta@eaton.com,
deepak.ajwani@ucd.ie
}

\begin{document}

\maketitle

\begin{abstract}
We propose a multi-stage learning approach for pruning the search space of maximum clique enumeration, a fundamental computationally difficult problem arising in various network analysis tasks. In each stage, our approach learns the characteristics of vertices in terms of various neighborhood features and leverage them to prune the set of vertices that are likely \emph{not} contained in any maximum clique. Furthermore, we demonstrate that our approach is domain independent -- the same small set of features works well on graph instances from different domain. Compared to the state-of-the-art heuristics and preprocessing strategies, the advantages of our approach are that (i) it does not require any estimate on the maximum clique size at runtime and (ii) we demonstrate it to be effective also for dense graphs. In particular, for dense graphs, we typically prune around 30 \% of the vertices resulting in speedups of up to 53 times for state-of-the-art solvers while generally preserving the size of the maximum clique (though some maximum cliques may be lost). For large real-world sparse graphs, we routinely prune over 99 \% of the vertices resulting in several tenfold speedups at best, typically with no impact on solution quality.
\end{abstract}

\section{Introduction}
A large number of optimization problems in diverse domains such as data mining, decision-making, planning, routing and scheduling are computationally hard (i.e., $\NP$-hard). No efficient polynomial-time algorithms are known for these problems that can solve every instance of the problem to optimality and many researchers consider that such algorithms may not even exist. A common way to deal with such optimization problems is to design heuristics that leverage the structure in real-world instance classes for these problems. This is a time-consuming process where algorithm engineers and domain experts have to identify the key characteristics of the instance classes and carefully design algorithm for optimality on instances with those characteristics.

In recent years, researchers have started exploring if machine learning techniques can be used to (i) automatically identify characteristics of the instance classes and (ii) learn algorithms specifically leveraging those characteristics. In particular, recent advances in deep learning and graph convolutional networks have been used in an attempt to directly \emph{learn} the output of an optimization algorithm based on small training examples (see e.g.,~\cite{Vinyals2015,Bello2016,Nowak2017}). These approaches have shown promising early results on some optimization problems such as the Travelling Salesman Problem (TSP). However, there are two fundamental challenges that limit the widespread adoption of these techniques: (i) requirement of large amounts of training data whose generation requires solving the $\NP$-hard optimization problem on numerous instances and (ii) the resultant lack of scalability (most of the reported results are on small test instances). 

Recently,~\cite{our-nips} proposed a probabilistic preprocessing framework to address the above challenges. Instead of directly learning the output of the $\NP$-hard optimization problem, their approach learns to prune away a part of the input. The reduced problem instance can then be solved with exact algorithms or constraint solvers. Because their approach merely needs to learn the elements of the input it can confidently prune away, it needs significantly less training. This also enables it to scale to larger test instances. They considered the problem of maximum clique enumeration and showed that on sparse real-world instances, their approach pruned 75-98\% of the vertices. Despite the conceptual novelty, the approach still suffered from (i) poor pruning on dense instances, (ii) poor accuracy on larger synthetic instances and (iii) non-transferability of training models across domains. In this paper, we build upon their work and show that we can achieve a significantly better accuracy-pruning trade-off, both on sparse and dense graphs, as well as cross-domain generalizability using a multi-stage learning methodology.

\textbf{Maximum clique enumeration}
We consider the maximum clique enumeration (MCE) problem, where the goal is to list all \emph{maximum} (as opposed to maximal) cliques in a given graph. The maximum clique problem is one of the most heavily-studied combinatorial problems arising in various domains such as in the analysis of social networks~\cite{soc,Fortunato2010,Palla2005,Papadopoulos2012}, behavioral networks~\cite{beha}, and financial networks~\cite{finan}. It is also relevant in clustering~\cite{dynamic,Yang2016} and cloud computing~\cite{Wang2014,Yao2013}. The listing variant of the problem, MCE, is encountered in computational biology~\cite{bio,Eblen2012,Yeger2004,mce} in problems like the detection of protein-protein interaction complex, clustering protein sequences, and searching for common cis-regulatory elements~\cite{protein}.

It is $\NP$-hard to even approximate the maximum clique problem within $n^{1-\epsilon}$ for any $\epsilon > 0$~\cite{Zuckerman2006}.
Furthermore, unless an unlikely collapse occurs in complexity theory, the problem of identifying if a graph of $n$ vertices has a clique of size $k$ is not solvable in time $f(k) n^{o(k)}$ for any function $f$~\cite{Chen2006}. As such, even small instances of this problem can be non-trivial to solve. Further, under reasonable complexity-theoretic assumptions, there is no polynomial-time algorithm that preprocesses an instance of $k$-clique to have only $f(k)$ vertices, where $f$ is any computable function depending solely on $k$ (see e.g.,~\cite{fpt-book}). These results indicate that it is unlikely that an efficient preprocessing method for MCE exists that can reduce the size of input instance drastically while guaranteeing to preserve all the maximum cliques. In particular, it is unlikely that polynomial-time sparsification methods (see e.g.,~\cite{Batson2013}) would be applicable to MCE. This has led researchers to focus on heuristic pruning approaches.


A typical preprocessing step in a state-of-the-art solver is the following: (i) quickly find a large clique (say of size $k$), (ii) compute the core number of each vertex of the input graph $G$, and (iii) delete every vertex of $G$ with core number less than $k-1$. This can be equivalently achieved by repeatedly removing all vertices with degree less than $k$. For example, the solver \texttt{pmc}~\cite{Rossi2015b} -- which is regarded as ``\emph{the} leading reference solver''~\cite{San2016} -- use this as the only preprocessing method. 
However, there are two major downsides to this preprocessing step.
First, it is crucially dependant on $k$, the size of a large clique found. 
Since the maximum clique size is $\NP$-hard to approximate within a factor of $n^{1-\epsilon}$, maximum clique estimates with no formal guarantees are used.
Second and more important, it is typical that even if the estimate $k$ was equal to the size of a maximum clique in $G$, the core number of most vertices could be considerably higher than $k-1$. This is particularly true in the case of dense graphs and it results in little or \emph{no} pruning of the search space. Similarly, other preprocessing strategies (see e.g.,~\cite{Eblen2010} for more discussion) depend on $\NP$-hard estimates of specific graph properties and are not useful for pruning dense graphs.


\paragraph{Our Results} 
We demonstrate 30 \% vertex pruning rates on average for dense networks, for which exact state-of-the-art methods are not able to prune anything, while typically only compromising the number of maximum cliques and not their size. For sparse networks, our preprocessor typically prunes well over 99 \% of the vertices without compromising the solution quality. In both cases, these prunings result in speedups as high as several tenfold for state-of-the-art MCE solvers. For example, after the execution of our multi-stage preprocessor, we correctly list all the~196 maximum cliques (of size~24) in a real-world social network (socfb-B-anon) with 3~M vertices and 21~M edges in only 7 seconds of solver time, compared with 40 minutes of solver time with current state-of-the-art preprocessor (see Table~\ref{tbl:pruning}).

\section{Preliminaries and Related Work}

Let $G=(V,E)$ be an undirected simple graph.
A \emph{clique} is a subset $S \subseteq V$ such that every two distinct vertices of $S$ are adjacent.
We say that the vertices of $S$ form a $k$-clique when $|S| = k$.
The \emph{clique number} of $G$, denoted by $\omega(G)$, is the size of a maximum clique in $G$.
A \emph{$k$-coloring} of $G$ is a function $c: V \to \{1, \ldots, k\}$. 
A \emph{coloring} is a $k$-coloring for some $k \le |V|$. 
A coloring $c$ is \emph{proper} if $c(u)\neq c(v)$ for every edge $\{u,v\} \in E$.
The \emph{chromatic number} of $G$, denoted by $\chi(G)$, is the smallest $k$ such that $G$ has a proper $k$-coloring.
It is easy to see that $\chi(G) \geq \omega(G)$ as at least $k$ colors are needed to color a $k$-clique.
Finally, a \emph{$k$-core} of a graph $G$ is a maximal subgraph of $G$ where every vertex in the subgraph has degree at least $k$ in the subgraph.
The \emph{core number} of a vertex $v$ is the largest $k$ for which a $k$-core containing $v$ exists.

\paragraph{Machine learning and $\NP$-hard problems}
There has been work on using machine learning to help tackle hard problems with different approaches. Some solve a problem by augmenting existing solvers~\cite{Liang2016}, predicting a suitable solver to run for a given instance~\cite{Fitzgerald2015,Loreggia2016}, or attempting to discover new algorithms~\cite{Khalil2017}. In contrast, some methods address the problems more directly. Examples include approaches to TSP~\cite{Hopfield1985,Fort1988,Durbin1987}, with recent work in~\cite{Vinyals2015,Bello2016,Nowak2017}.

\paragraph{Maximal clique enumeration} We note that there are algorithms~\cite{Eppstein2010,Cheng2011} for \emph{maximal} clique enumeration, in contrast to our problem of \emph{maximum} clique enumeration. The two set of algorithms are required in very different applications, and the runtime of maximal clique enumeration is generally significantly higher. 

\paragraph{Probabilistic preprocessing}
Recently,~\cite{our-nips} proposed a probabilistic preprocessing framework for fine-grained search space classification.
It treats individual vertices of $G=(V,E)$ as classification problems and the problem of learning a preprocessor reduces to that of learning a mapping $\gamma : V \to \{0,1\} $ from a set of $L$ training examples $T = \{ \langle f(v_i), y_i \rangle \}^L_{i=1}$, where $v_i \in V$ is a vertex, $y_i \in \{0,1\}$ a class label, and $f : V \to \mathbb{R}^d$ a mapping from a vertex to a $d$-dimensional feature space. To learn the mapping $\gamma$ from $T$, a probabilistic classifier $P$ is used which outputs a probability distribution over $\{0,1\}$ for a given $f(u)$ for $u \in V$. 
Then, on input graph $G$, all vertices from $G$ that are predicted by $P$ to not be in a solution with probability at least $q$ (for some \emph{confidence threshold} $q$) are pruned away. 
Here, $q$ trades-off the pruning rate with the accuracy of the pruning. 

This framework showed that there is potential for learning a heuristic preprocessor for instance size pruning. However, the speedups obtained were limited and the training models were not transferable across domains. We build upon this work and show that we can achieve cross-domain generalizability and considerable speedups, both on sparse and dense graphs, using a multi-stage learning methodology.

\section{Proposed framework}
\label{sec:framework}
In this section, we introduce our multi-stage preprocessing approach and then give the features that we use for pruning.

\paragraph{Multi-stage sparsification}
A major difficulty with the probabilistic preprocessing described above is that when training on sparse graphs, the learnt model focused too heavily on pruning out the easy cases, such as low-degree vertices and not on the difficult cases like vertices with high degree and high core number. To improve the accuracy on difficult vertices, we propose a multi-stage sparsification approach. In each stage, the approach focuses on gradually harder cases that were difficult to prune by the classifier in earlier stages.

Let $\mathcal{G}_1$ be the input set of networks.
Consider a graph $G \in \mathcal{G}_1$. 
Let $\mathcal{M}$ be the set of all maximum cliques of $G$, and denote by $V(\mathcal{M})$ the set of all vertices in $\mathcal{M}$.
The positive examples in the training set $T_1$ consist of all vertices that are in some maximum clique ($V(\mathcal{M})$) and the negative examples are the ones in the set $V \setminus V(\mathcal{M})$. 
Since the training dataset can be highly skewed, we under-sample the larger class to achieve a balanced training data. A probabilistic classifier $P_1$ is trained on the balanced training data in stage $1$. Then, in the next stage, we remove all vertices that were predicted by $P_1$ to be in the negative class with a probability above a predefined threshold $q$. We focus on the set $\mathcal{G}_2$ of subgraphs (of graphs in $\mathcal{G}_1$) induced on the remaining vertices and repeat the above process. The positive examples in the training set $T_2$ consists of all vertices in some maximum clique ($V(\mathcal{M})$) and the negative examples are the ones in the set $V \setminus V(\mathcal{M})$, training dataset is balanced by under-sampling and we use that balanced dataset to learn the probabilistic classifier $P_2$. We repeat the process for $\ell$ stages.

As we show later, the multi-stage sparsification results in significantly more pruning compared to a single-stage probabilistic classifier. 

\begin{figure}[t]
    \centering
        \includegraphics[width=0.30\textwidth,keepaspectratio]{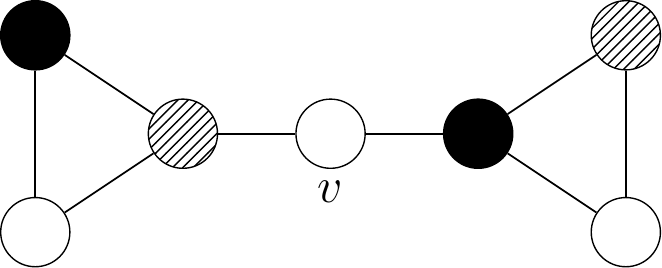} 
    \caption{While the shown proper 3-coloring is optimal, we can swap the non-white colors in either triangle to see that $\chi_d(v) = 1/3$.}
    \label{fig:chrom-density}
\end{figure}

\paragraph{Graph-theoretic features}
We use the following graph-theoretic features: \textbf{(F1)} number of vertices, \textbf{(F2)} number of edges, \textbf{(F3)} vertex degree, \textbf{(F4)} local clustering coefficient (LCC), and \textbf{(F5)} eigencentrality.

The crude information captured by features (F1)-(F3) provide a reference for the classifier for generalizing to different distributions from which the graph might have been generated. 
Feature (F4), the LCC of a vertex is the fraction of its neighbors with which the vertex forms a triangle, encapsulating the well-known small world phenomenon.  
Feature (F5) eigencentrality represents a high degree of connectivity of a vertex to other vertices, which in turn have high degrees as well. 
The \emph{eigenvector centrality} $\vec{v}$ is the eigenvector of the adjacency matrix $A$ of $G$ with the largest eigenvalue $\lambda$, i.e., it is the  solution of $\vec{A}\vec{v} = \lambda\vec{v}$.
The $i$th entry of $\vec{v}$ is the \emph{eigencentrality} of vertex $v$. In other words, this feature provides a measure of local ``denseness''. A vertex in a dense region shows higher probability of being part of a large clique.

\paragraph{Statistical features}
In addition, we use the following statistical features: \textbf{(F6)} the $\chi^2$ value over vertex degree, \textbf{(F7)} average $\chi^2$ value over neighbor degrees, \textbf{(F8)} $\chi^2$ value over LCC, and \textbf{(F9)} average $\chi^2$ value over neighbor LCCs.

The intuition behind (F6)-(F9) is that for a vertex $v$ present in a large clique, its degree and LCC would deviate 
from the underlying expected distribution characterizing the graph. 
Further, the neighbors of $v$ also present in the clique would demonstrate such behaviour.
Indeed, statistical features have been shown to be robust in approximately capturing local structural patterns~\cite{graph}.

Statistical significance is captured by the notion of p-value~\cite{fitStatistics}, and well-estimated~\cite{pear} by the {\em Pearson's chi-square statistic}, $\chi^2$, computed as $\chi^2 = \sum_{\forall i}\left[\left(O_i - E_i\right)^2 / E_i\right]$,
where $O_i$ and $E_i$ are the observed and expected number of occurrences of the possible outcomes $i$.


\paragraph{Local chromatic density}
Let $G=(V,E)$ be a graph.
We define the \emph{local chromatic density} of a vertex $v \in V$, denoted by $\chi_d(v)$, as the minimum ratio of the number of distinct colors appearing in $N(v)$ and any optimal proper coloring of $G$.
Put differently, the local chromatic density of $v$ is the minimum possible number of colors in the immediate neighborhood of $v$ in any optimal proper coloring of $G$ (see Figure~\ref{fig:chrom-density}).

We use the local chromatic density as the feature \textbf{(F10)}.
A vertex $v$ with high $\chi_d(v)$ means that the neighborhood of $v$ is dense, as it captures the adjacency relations between the vertices in $N(v)$.
Thus, a vertex in such a dense region has a higher chance of belonging to a large clique.

However, the problem of computing $\chi_d(v)$ is computationally difficult.
In the decision variant of the problem, we are given a graph $G=(V,E)$, a vertex $v \in V$, and a ratio $q \in (0,1)$. 
The task is to decide whether there is proper $k$-coloring $c$ of $V$ witnessing $\chi_d(v) \geq q$.
The omitted proof is by a polynomial-time reduction from graph coloring.
\begin{theorem}
Given a graph $G=(V,E)$, $v \in V$, and $q \in (0,1)$, it is $\NP$-hard to decide whether $\chi_d(v) \leq q$.
\end{theorem}
%
\noindent Despite its computational hardness, we can in practice compute $\chi_d(v)$ by a heuristic.
Indeed, to compute $\chi_d(v)$ for every $v \in V$, we first compute a proper coloring for $G$ using e.g., the well-known linear-time greedy heuristic of~\cite{Welsh1967}.
After a proper coloring has been computed, we compute the described ratio for every vertex from that.

\begin{table}[t]
  \centering
  \small
  \caption{The effect of introducing the feature (F10) the local chromatic density into the feature set. The column ``W/o'' is the vertex classification accuracy of the classifier of~[Lauri and Dutta, 2019] without (F10) while column ``With'' is the same with (F10).}
  \label{tbl:chrom-feat}
  \def\arraystretch{1}
  \setlength{\tabcolsep}{3pt}
  \begin{tabular}{*{111}{l}}
    \toprule
    \multicolumn{2}{c}{\textbf{bio}} & \multicolumn{2}{c}{\textbf{soc}} & \multicolumn{2}{c}{\textbf{socfb}} & \multicolumn{2}{c}{\textbf{web}} & \multicolumn{2}{c}{\textbf{all}} \\
    \cmidrule(lr){1-2}
    \cmidrule(lr){3-4}
    \cmidrule(lr){5-6}
    \cmidrule(lr){7-8}
    \cmidrule(lr){9-10}
    \textbf{W/o} & \textbf{With} & \textbf{W/o} & \textbf{With} & \textbf{W/o} & \textbf{With} & \textbf{W/o} & \textbf{With} & \textbf{W/o} & \textbf{With} \\
    0.95   & 0.98    & 0.89   & 0.99   & 0.90   & 0.95    & 0.96   & 0.99    & 0.87  & 0.96    \\
    \bottomrule
  \end{tabular}
\end{table}

\paragraph{Learning over edges}
Instead of individual vertices, we can view the framework also over \emph{individual edges}.
In this case, the goal is to find a mapping $\gamma' : E \to \{0,1\}$, and the training set $L'$ contains feature vectors corresponding to edges instead of vertices.
We also briefly explore this direction in this work.

\paragraph{Edge features}
We use the following features (E1)-(E9) for an edge $\{u,v\}$.
\textbf{(E1)} Jaccard similarity is the number of common neighbors of $u$ and $v$  divided by the number of vertices that are neighbors of at least one of $u$ and $v$. 
\textbf{(E2)} Dice similarity is twice the number of common neighbors of $u$ and $v$, divided by the sum of their degrees.
\textbf{(E3)} Inverse log-weighted similarity is as the number of common neighbors of $u$ and $v$ weighted by the inverse logarithm of their degrees.
\textbf{(E4)} Cosine similarity is the number of common neighbors of $u$ and $v$ divided by the geometric mean of their degrees.
The next three features are inspired by the vertex features: \textbf{(E5)} average LCC over $u$ and $v$, \textbf{(E6)} average degree over $u$ and $v$, and \textbf{(E7)} average eigencentrality over $u$ and $v$.
\textbf{(E8)} is the number of length-two paths between $u$ and $v$.
Finally, we use \textbf{(E9)} \emph{local edge-chromatic density}, i.e., the number of distinct colors on the common neighbors of $u$ and $v$ divided by the total number of colors used in any optimal proper coloring.

The intuition behind (E1)-(E4) is well-established for community detection; see e.g.,~\cite{Harenberg2014} for more.
For (E8), observe that the number of length-two paths is high when the edge is part of a large clique, and at most $n-2$ when $\{u,v\}$ is an edge of a complete graph on $n$ vertices.
Notice that (E9) could be converted into a deterministic rule: the edge $\{u,v\}$ can be safely deleted if the common neighbors of $u$ and $v$ see less than $k-2$ colors in any proper coloring of the input graph $G$, where $k$ is an estimate for $\omega(G)$.
To our best knowledge, such a rule has not been considered previously in the literature.
Further, notice that there are situations in which this rule \emph{can} be applied whereas the similar vertex rule uncovered from (F10) cannot.
To see this, let $G$ be a graph consisting of two triangles $\{a,b,c\}$ and $\{x,y,z\}$, connected by an edge $\{a,x\}$, and let $k = 3$.
The vertex rule cannot delete $a$ nor $x$, but the described edge rule removes $\{a,x\}$.

\section{Experimental results}
In this section, we describe how multi-stage sparsification is applied to the MCE problem and our computational results.

To allow for a clear comparison, we follow closely the definitions and practices specified in~\cite{our-nips}. Thus, unless otherwise mentioned and to save space, we refer the reader to that work for additional details.

All experiments ran on a machine with Intel Core i7-4770K CPU (3.5 GHz), 8 GB of RAM, running Ubuntu 16.04.

\paragraph{Training and test data}
All our datasets are obtained from Network Repository~\cite{Rossi2015} (available at \url{http://networkrepository.com/}).

For dense networks, we choose a total of 30 networks from various categories with the criteria that the edge density is at least 0.5 in each.
We name this category ``dense''.
The test instances are in Table~\ref{tbl:dense}, chosen based on empirical hardness (i.e., they are solvable in reasonable amount of time).

For sparse networks, we choose our training data from four different categories: 31 biological networks (``bio''), 32 social networks (``soc''), 107 Facebook networks (``socfb''), and 13 web networks (``web'').
In addition, we build a fifth category ``all'' that comprises all networks from the mentioned four categories.
The test instances are in Table~\ref{tbl:pruning}.

\paragraph{Feature computation}
We implement the feature computation in C++, relying on the \texttt{igraph}~\cite{igraph} C graph library.
In particular, our feature computation is single-threaded with further optimization possible.


\paragraph{Domain oblivious training via local chromatic density}
In~\cite{our-nips}, it was assumed that the classifier should be trained with networks coming from the same domain, and that testing should be performed on networks from that domain.
However, we demonstrate in Table~\ref{tbl:chrom-feat} that a classifier can be trained with networks from various domains, yet predictions remain accurate across domains (see column ``all'').
The accuracy is boosted considerably by the introduction of the local chromatic density (F10) into the feature set (see Table~\ref{tbl:chrom-feat}).
In particular, when generalizing across various domains, the impact on accuracy is almost 10~\%. 
For this reason, rather than focusing on network categories, we only consider networks by edge density (at least 0.5 or not).

\paragraph{State-of-the-art solvers for MCE}
To our best knowledge, the only publicly available solvers able to list all maximum cliques\footnote{For instance, \texttt{pmc}~\cite{Rossi2015b} does not have this feature.} are
\texttt{cliquer}~\cite{Ostergard2002}, based on a branch-and-bound strategy; and \texttt{MoMC}~\cite{Li2017}, introducing incremental maximum satisfiability reasoning to a branch-and-bound strategy. 
We use these solvers in our experiments.

\subsection{Dense networks}
\label{subs:dense}
In this subsection, we show results for probabilistic preprocessing on dense networks (i.e., edge density at least 0.5).

\begin{table*}[t]
\centering
\small
\caption{Experiments for dense graphs. The column ``$\omega$'' is the max.\ clique size and the column ``n.\ $\omega$'' is the number of such cliques. In both, * means the quantity is preserved in the preprocessed instance; otherwise the new quantity is in parenthesis.
The multicolumns ``$k$-core'' and ``1-stage'' give the vertex pruning ratio followed by the edge pruning ratio when preprocessed by removing vertices of core number $< \omega - 1$ and our preprocessor, respectively.
For the last three columns, all runtimes are in seconds averaged over three independent runs. The column ``Pruning'' is the time for feature computation \emph{and} pruning. The two remaining columns give the runtime of a solver, containing the runtime on the pruned instance with the speedup obtained in parenthesis. We denote by \texttt{t/o} killed execution after an hour and --- denotes no speedup.}
\def\arraystretch{0.95}
\label{tbl:dense}
\begin{tabular}{lrrrrrrrrrrr}
\toprule
Instance & $|V|$ & $|E|$ & $\omega$ & n.\ $\omega$ & \multicolumn{2}{c}{$k$-core} & \multicolumn{2}{c}{1-stage} & Pruning & \texttt{cliquer} & \texttt{MoMC} \\
\midrule
brock200-1 & 200 & 14.8 K & 21 (20) & 2 (16) & --- & --- & \textbf{0.34} & \textbf{0.55} & $<$0.01 & \textbf{0.39 (53.07)}& 0.04 (44.57) \\
keller4 & 171 & 9.4 K & \textbf{11*} & 2304 (37) & --- & --- & \textbf{0.30} & \textbf{0.50} & $<$0.01 & \textbf{$<$0.01 (38.11)} & 0.02 (5.68) \\
\rowcol keller5 & 776 & 226 K & \textbf{27*} & 1000 (5) & --- & --- & \textbf{0.28} & \textbf{0.48} & 0.19 & \texttt{t/o} & \textbf{1421.24 ($>$2.53)} \\
p-hat300-3 & 300 & 33.4 K & \textbf{36*} & \textbf{10*} & --- & --- & \textbf{0.38} & \textbf{0.58} & 0.02 & \textbf{87.1 (9.12)} & 0.05 (6.00) \\
p-hat500-3 & 500 & 93.8 K & \textbf{50*} & 62 (40) & --- & --- & \textbf{0.34} & \textbf{0.52} & 0.07 & \texttt{t/o} & \textbf{2.51 (5.98)} \\
\rowcol p-hat700-1 & 700 & 61 K & \textbf{11*} & \textbf{2*} & --- & --- & \textbf{0.36} & \textbf{0.47}  & 0.03 & 0.08 (1.22) & \textbf{0.05 (1.30)} \\
p-hat700-2 & 700 & 121.7 K & \textbf{44*} & \textbf{138*} & --- & --- & \textbf{0.36} & \textbf{0.45} & 0.11 & \texttt{t/o} & 1.35 (---)  \\
p-hat1000-1 & 1 K & 122.3 K & \textbf{10*} & 276 (165) & --- & --- & \textbf{0.36} & \textbf{0.47}  & 0.08 & \textbf{0.86 (2.22)} & 0.71 (1.67) \\
\rowcol p-hat1500-1 & 1.5 K & 284.9 K & 12 (11) & 1 (376) & --- & --- & \textbf{0.33} & \textbf{0.43} & 0.25 & 13.18 (---) & \textbf{3.2 (1.54)} \\
fp & 7.5 K & 841 K & \textbf{10*} & \textbf{1001*} & --- & --- & \textbf{0.06} & \textbf{0.29} & 0.36 & 0.65 (---) & \textbf{5.19 (1.13)} \\
nd3k & 9 K & 1.64 M & \textbf{70*} & \textbf{720*} & --- & --- & \textbf{0.23} & \textbf{0.28} & 1.28 & \texttt{t/o} & \textbf{7.05 (1.09)} \\
\rowcol raefsky1 & 3.2 K & 291 K & \textbf{32*} & 613 (362) & --- & --- & \textbf{0.33} & \textbf{0.38} & 0.11 & 2.80 (---) & \textbf{0.31 (1.36)} \\
HFE18\_96\_in & 4 K & 993.3 K & \textbf{20*} & \textbf{2*} & $<$1e-4 & $<$1e-4 & \textbf{0.26} & \textbf{0.27} & 0.49 & 58.88 (1.05) & \textbf{4.30 (1.18)} \\
heart1 & 3.6 K & 1.4 M & \textbf{200*} & 45 (26) & $<$1e-4 & $<$1e-4 & \textbf{0.19} & \textbf{0.25} & 0.66 & \texttt{t/o} & 19.37 (---) \\
\rowcol cegb2802 & 2.8 K & 137.3 K & \textbf{60*} & 101 (38) & 0.09 & 0.04 & \textbf{0.39} & \textbf{0.46} & 0.09 & 0.05 (---) & \textbf{0.15 (1.61)} \\
movielens-1m & 6 K & 1 M & \textbf{31*} & \textbf{147*} & 0.05 & 0.007 & \textbf{0.22} & \textbf{0.23} & 0.98 & 31.31 (---) & \textbf{2.85 (1.14)} \\
ex7 & 1.6 K & 52.9 K & \textbf{18*} & 199 (127) & 0.02 & 0.01 & \textbf{0.26} & \textbf{0.28}  & 0.04 & 0.01 (---) & \textbf{0.1 (1.29)} \\
\rowcol Trec14 & 15.9 K & 2.87 M & \textbf{16*} & \textbf{99*} & 0.16 & 0.009 & \textbf{0.34} & \textbf{0.15} & 2.19 & 3.62 (---) & 0.35 (---) \\
\bottomrule 
\end{tabular}
\end{table*}

\paragraph{Classification framework for dense networks}

For training, we get 4762 feature vectors from our ``dense'' category.
As a baseline, a 4-fold cross validation over this using logistic regression from~\cite{our-nips} results in an accuracy of \textbf{0.73}.
We improve on this by obtaining an accuracy of \textbf{0.81} with gradient boosted trees (further details omitted), found with the help of \texttt{auto-sklearn}~\cite{autosklearn}.

\paragraph{Search strategies}
Given the empirical hardness of dense instances, one should not expect a very high accuracy with polynomial-time computable features such as (F1)-(F10).
For this reason, we set the confidence threshold $q=0.98$ here.


\paragraph{The failure of $k$-core decomposition on dense graphs}
It is common that widely-adopted preprocessing methods like the $k$-core decomposition cannot prune any vertices on a dense network $G$, even if they had the computationally expensive knowledge of $\omega(G)$.
This is so because the degree of each vertex is higher than than the maximum clique size $\omega(G)$.

We showcase precisely this poor behaviour in Table~\ref{tbl:dense}.
For most of the instances, the $k$-core decomposition with the exact knowledge of $\omega(G)$ cannot prune any vertices.
In contrast, the probabilistic preprocessor prunes typically around 30 \% of the vertices and around 40 \% of the edges.

\paragraph{Accuracy}
Given that around 30 \% of the vertices are removed, how many mistakes do we make?
For almost all instances we retain the clique number, i.e., $\omega(G') = \omega(G)$, where $G'$ is the instance obtained by preprocessing $G$ (see column ``$\omega$'' in Table~\ref{tbl:dense}).
In fact, the only exceptions are \mbox{brock200-1} and \mbox{p-hat1500-1}, for which $\omega(G') = \omega(G) - 1$ still holds.
Importantly, for about half of the instances, we retain \emph{all} optimal solutions.

\paragraph{Speedups}
We show speedups for the solvers after executing our pruning strategy in Table~\ref{tbl:dense} (last two columns).
We obtain speedups as large as 53x and for 38x \mbox{brock200-1} and \mbox{keller4}, respectively.
This might not be surprising, since in both cases we lose some maximum cliques (but note that for \mbox{keller4}, the size of a maximum clique is still retained).
For \mbox{p-hat300-3}, the preprocessor makes no mistakes, resulting in speedups of upto 9x.
The speedup for \mbox{keller5} is \emph{at least} 2.5x, since the original instance was not solved within 3600 seconds, but the preprocessed instances was solved in roughly 1421 seconds.

Most speedups are less than 2x, explained by the relative simplicity of instances.
Indeed, it seems challenging to locate dense instances of MCE that are (i) structured and (ii) solvable within a reasonable time.


\subsection{Sparse networks}
\label{subs:sparse}
In this subsection, we show results for probabilistic preprocessing on sparse networks (i.e., edge density below 0.5).

\paragraph{Classification framework for sparse networks} We use logistic regression trained with stochastic gradient descent.

\begin{table*}[t]
\centering
\small
\caption{Experiments for sparse graphs. The columns are precisely as in Table~\ref{tbl:dense}, with the exception that we show pruning ratios for 5 stages.
All ratios are rounded to three decimal places. 
Underlined ratios of 1.000 mean the ratio is precisely 1, otherwise it is between 1 and 0.999.}
\def\arraystretch{0.95}
\label{tbl:pruning}
\begin{tabular}{lrrrrrrrrrrr}
\toprule
Instance & $|V|$ & $|E|$ & $\omega$ & n.~$\omega$ & \multicolumn{2}{c}{$k$-core} & \multicolumn{2}{c}{5-stage} & Pruning & \texttt{cliquer} & \texttt{MoMC} \\
\midrule
bio-WormNet-v3 & 16 K & 763 K &  \textbf{121*} & \textbf{18*} & 0.868 & 0.602 & \textbf{0.987} & \textbf{0.975} & 0.36 & 0.37 (---) & \textbf{0.40 (3.94)} \\
ia-wiki-user-edits-page & 2 M & 9 M & \textbf{15*} & \textbf{15*} & 0.958 & 0.641 & \textbf{0.997}	& \textbf{0.946} & 1.12 & \textbf{1.16 (29.94)} & \texttt{s} \\
\rowcol rt-retweet-crawl & 1 M & 2 M &   \textbf{13*} &   \textbf{26*} &      0.979 & 0.863 & \textbf{0.997} & \textbf{0.989} & 0.38 & \textbf{0.41 (5.66)} & \texttt{s} \\
soc-digg	& 771 K & 6 M & \textbf{50*} & \textbf{192*} & 0.969 & 0.496 & \textbf{0.998}	& \textbf{0.964} & 4.80 & \textbf{4.91 (1.78)} & \texttt{s} \\
soc-flixster & 3 M & 8 M &    \textbf{31*} &  \textbf{752*}  &      0.986 & 0.834 & \textbf{0.999} & \textbf{0.989} & 1.32 & \textbf{1.41 (3.86)} & \texttt{s} \\
\rowcol soc-google-plus & 211 K & 2 M &   \textbf{66*} &   \textbf{24*} &      0.986 & 0.785 & \textbf{0.998} & \textbf{0.972} & 0.35 & 0.35 (---) & \textbf{0.41 (3.98)} \\
soc-lastfm & 1 M & 5 M &    \textbf{14*} &  330 (324) &      0.933 & 0.625 & \textbf{0.993} & \textbf{0.938} & 2.24 & \textbf{2.57 (10.56)} & \texttt{s} \\
soc-pokec & 2 M & 22 M & \textbf{29*} & \textbf{6*} & 0.824 & 0.595 & \textbf{0.975} & \textbf{0.940} & 17.59 & \textbf{24.40 (45.80)} & \texttt{s} \\
\rowcol      soc-themarker & 69 K & 2 M &   \textbf{22*} &   \textbf{40*} &      0.713 & 0.151 & \textbf{0.972}	& \textbf{0.842} & 2.03 & \textbf{4.95 (---)} & \texttt{s} \\
soc-twitter-higgs	& 457 K & 15 M & \textbf{71*} & \textbf{14*} & 0.852 & 0.540 & \textbf{0.986}	& \textbf{0.943} & 9.52 & \textbf{9.85 (1.92)} & \texttt{s} \\
 soc-wiki-Talk-dir & 2 M & 5 M &   \textbf{26*} &  \textbf{141*} &      0.993 & 0.830 & \textbf{0.999} & \textbf{0.970} & 1.09 & \textbf{3.47 (1.25)} & \texttt{s} \\
\rowcol      socfb-A-anon & 3 M & 24 M &   \textbf{25*} &   \textbf{35*} &      0.879 & 0.403 & \textbf{0.984} & \textbf{0.907} & 28.49 & \textbf{38.05 (55.95)} & \texttt{s} \\
      socfb-B-anon & 3 M & 21 M &   \textbf{24*} &  \textbf{196*} &     0.884 & 0.378 & \textbf{0.986} & \textbf{0.920} & 28.33 & \textbf{35.49 (67.46)} & \texttt{s} \\
     socfb-Texas84 & 36 K & 2 M &  \textbf{51*} &   \textbf{34*} &      0.540 & 0.322 & \textbf{0.957} & \textbf{0.941} & 1.04 & \textbf{1.07 (1.32)} & \texttt{s} \\
\rowcol tech-as-skitter & 2 M & 11 M &   \textbf{67*} &    \textbf{4*}  &      0.997 & 0.971 & \textbf{1.000} & \textbf{0.998} & 0.28 & 0.28 (---) & \textbf{0.36 (4.31)} \\
   web-baidu-baike & 2 M & 18 M & \textbf{31*} & \textbf{4*} & 0.933 & 0.618 & \textbf{0.992} & \textbf{0.934} & 9.67 & \textbf{11.00 (7.48)} & \texttt{s} \\
    web-google-dir & 876 K & 5 M &   \textbf{44*} &    \textbf{8*} &      1.000 & 0.999 & \textbf{1.000} & \textbf{1.000} & $<$ 0.00 & $<$ 0.00 (---) & \textbf{$<$ 0.00 (2.06)} \\
\rowcol        web-hudong & 2 M & 15 M &  267 (266) &   59 (1) &      1.000 & 0.996 & \textbf{1.000} & \textbf{0.997} & 0.09 & 0.10 (---) & \textbf{0.1 (9.99)} \\
 web-wikipedia2009 & 2 M & 5 M &   \textbf{31*} &    \textbf{3*} & 0.999 & 0.988 & \textbf{1.000} & \textbf{1.000} & 0.03 & 0.03 (---) & \textbf{0.03 (4.28)} \\
\bottomrule
\end{tabular}
\end{table*}

\paragraph{Implementing the $k$-core decomposition}
Recall the exact state-of-the-art preprocessor: (i) use a heuristic to find a large clique (say of size $k$) and (ii) delete every vertex of $G$ of core number less than $k-1$.
For sparse graphs, a state-of-the-art solver \texttt{pmc} has been reported to find large cliques, i.e., typically $k$ is at most a small additive constant away from $\omega(G)$ (a table of results seen at \url{http://ryanrossi.com/pmc/download.php}).
Further, given that some real-world sparse networks are scale-free (many vertices have low degree) the $k$-core decomposition can be effective in practice.

To ensure highest possible prune ratios for the $k$-core decomposition method, we supply it with the number $\omega(G)$ instead of an estimate provided by any real-world implementation.
This ensures \emph{ideal conditions}: (i) the method always prunes as aggressively as possible, and (ii) we further assume its execution has zero cost.
We call this method the \emph{$\omega$-oracle}.

\paragraph{Test instance pruning}
Before applying our preprocessor on the sparse test instances, we prune them using the $\omega$-oracle.
This ensures that the pruning we report is highly non-trivial, while also speeding up feature computation.

\paragraph{Search strategies}
We experiment with the following two multi-stage search strategies:
\begin{itemize}
\item \emph{Constant confidence (CC):} at every stage, perform probabilistic preprocessing with confidence threshold $q$.
\item \emph{Increasing confidence (IC):} at the first stage, perform probabilistic preprocessing with confidence threshold $q$, progressing $q$ by $d$ for every later stage.
\end{itemize}
Our goal is two-fold: to find (i) a number of stages $\ell$ and (ii) parameters $q$ and $d$, such that the strategy never errs while pruning as aggressively as possible.
We do a systematic search over parameters $\ell$, $q$, and $d$. 
For the CC strategy, we let $\ell \in \{1,2,\ldots,8\}$ and $q \in \{ 0.55, 0.6, \ldots, 0.95 \}$.
For the IC strategy, we try $q \in \{ 0.55, 0.60, 0.65 \}$, $d = 0.05$, and set $\ell$ so that in the last stage the confidence is 0.95.

We find the CC strategy with $q = 0.95$ to prune the highest while still retaining all optimal solutions.
Thus, for the remaining experiments, we use a CC strategy with $q=0.95$.

Our 5-stage strategy outperforms, almost always safely, the $\omega$-oracle (see Table~\ref{tbl:pruning}).
In particular, note that even if the difference between the vertex pruning ratios is small, the impact for the number of edges removed can be considerable (see e.g., all instances of the ``soc'' category).


\paragraph{Speedups}
We show speedups for the solvers in Table~\ref{tbl:pruning}.
We use as a baseline the solver executed on an instance pruned by the $\omega$-oracle, which renders many of the instances easy already.
Most notably, this is \emph{not} the case for \mbox{soc-pokec}, \mbox{socfb-A-anon}, and \mbox{socfb-B-anon}, all requiring at least 5 minutes of solver time.
The largest speedup is for \mbox{socfb-B-anon}, where we go from requiring 40 minutes to only 7 seconds of solver time.
For \texttt{MoMC}, most instances report a segmentation fault for an unknown reason.

\paragraph{Comparison against Lauri and Dutta}
The results in Table~\ref{tbl:pruning} are not directly comparable to those in~\cite[Table~1]{our-nips}.
First, the authors only give vertex pruning ratios.
While the difference in vertex pruning ratios might sometimes seem underwhelming, even small increases can translate to large decrements in the number of edges.
On the other hand, the difference is often clear in our favor as in \mbox{socfb-Texas84} and \mbox{bio-WormNet-v3} (i.e., 0.76 vs.\ 0.96 and 0.90 vs.\ 0.99).
Second, the authors use estimates on $\omega(G)$ -- almost always less than the exact value -- whereas we use the exact value provided by the $\omega$-oracle.
Thus, the speedups we report are \emph{as conservative as possible} unlike theirs.

\subsection{Edge-based classification}
For edges, we do a similar training as that described for vertices.
For the category ``dense'', we obtain 79472 feature vectors.
Further, for this category, the edge classification accuracy is \textbf{0.83}, which is 1~\% higher than the vertex classification accuracy using the same classifier as in Subsection~\ref{subs:dense}. However, we note that the edge feature computation is noticeably slower than that for vertex features.

\subsection{Model analysis}

Gradient boosted trees (used with dense networks in Subsection~\ref{subs:dense}) naturally output feature importances.
We apply the same classifier for the sparse case to allow for a comparison of feature importance.
In both cases, the importance values are distributed among the ten features and sum up to one.


Unsurprisingly, for sparse networks, the local chromatic density (F10) dominates (importance 0.22).
In contrast, (F10) is ineffective for dense networks (importance 0.08), since the chromatic number tends to be much higher than the maximum clique size.
In both cases, (F5) eigencentrality has relatively high importance, justifying its expensive computation.

For dense networks, (F7) average $\chi^2$ over neighbor degrees has the highest importance (importance 0.23), whereas in the sparse case it is least important feature (importance 0.03).
This is so because all degrees in a dense graph are high and the degree distribution tends to be tightly bound or coupled.
Hence, even slight deviations from the expected (e.g., vertices in large cliques) depict high statistical significance scores.

\section{Discussion and conclusions}
\label{sec:disc}
We proposed a multi-stage learning approach for pruning the search space of MCE, generalizing an earlier framework of~\cite{our-nips}. In contrast to known exact preprocessing methods, our approach requires no estimate for the maximum clique size at runtime -- a task $\NP$-hard to even approximate -- and particularly challenging on dense networks. We provide an extensive empirical study to show that our approach can routinely prune over 99~\% of vertices in sparse graphs. More importantly, our approach can typically prune around 30~\% of the vertices on dense graphs, which is considerably more than the existing methods based on $k$-cores.

\paragraph{Future improvements}
To achieve even larger speedups, one can consider parallelization of the feature computation (indeed, our current program is single-threaded).
In addition, at every stage, we recompute all features from scratch.
There are two obvious ways to speed this part: (i) it is unnecessary to recompute a local feature (e.g., degree or local clustering coefficient) for vertex $v$ if none of its neighbors were removed, and (ii) more generally, there is considerable work in the area of dynamic graph algorithms under vertex deletions.
Another improvement could be to switch more accurate but expensive methods for feature computation (e.g., (F10) which is $\NP$-hard) when the graph gets small enough.

\paragraph{Dynamic stopping criteria}
We refrained from multiple stages of preprocessing for dense networks due to the practical hardness of the task.
However, for sparse networks, it was practically always safe to perform five (or even more) of stages of preprocessing with no effect on solution quality.
An intriguing open problem is to propose a dynamic strategy for choosing a suitable number of stages $\ell$.
There are several possibilities, such as stopping when pruning less than some specified threshold, or always pruning aggressively (say up to $\ell=10$), computing a solution, and then backtracking by restoring the vertices deleted in the previous stage, halting when the solution does not improve anymore.

\paragraph{Classification over edges}
To speed up our current implementation for edge feature computation, a first step could be a well-engineered neighborhood intersection to speed up (E1)-(E4).
Luckily, this is a core operation in database systems with many high-performance implementations available~\cite{Lemire2016,Inoue2014,Lemire2015}.
Further path-based features are also possible, like the number of length-$d$ paths for $d=3$, which is still computed cheaply via e.g., matrix multiplication.
For larger $d$, one could rely on estimates based on random walk sampling.
In addition, it is possible to leave edge classification for the later stages, once the vertex classifier has reduced the size of the input graph enough.
We believe that there is further potential in exploring this direction.

\bibliographystyle{named}
\bibliography{ijcai19}

\end{document}